\documentclass[conference]{IEEEtran}
\IEEEoverridecommandlockouts
\usepackage{cite}
\usepackage{amsmath,amssymb,amsfonts}
\usepackage{latexsym}
\usepackage{algorithmic}
\usepackage[dvipdfmx]{graphicx}
\usepackage{textcomp}
\usepackage[dvipdfmx]{color}

\newcommand{\tcr}[1]{\textcolor{red}{#1}}

\newcommand{\ctext}[1]{\raise0.2ex\hbox{\textcircled{\scriptsize{#1}}}}

\def\BibTeX{{\rm B\kern-.05em{\sc i\kern-.025em b}\kern-.08em
    T\kern-.1667em\lower.7ex\hbox{E}\kern-.125emX}}
\begin{document}

\title{Translating MFM into FOL:\break towards plant operation planning}

\author{\IEEEauthorblockN{Shota Motoura}
\IEEEauthorblockA{\textit{Central Research Laboratories} \\
\textit{NEC Corporation}\\
Kanagawa, Japan \\
motoura@ct.jp.nec.com}
\and
\IEEEauthorblockN{Kazeto Yamamoto}
\IEEEauthorblockA{\textit{Central Research Laboratories} \\
\textit{\tcr{NEC Corporation AISTも追加}}\\
Kanagawa, Japan \\
k-yamamoto@ft.jp.nec.com}
\and
\IEEEauthorblockN{Shumpei Kubosawa}
\IEEEauthorblockA{\textit{Central Research Laboratories} \\
\textit{\tcr{NEC Corporation AISTも追加}}\\
Kanagawa, Japan \\
kubosawa@cd.jp.nec.com}
\and
\IEEEauthorblockN{Takashi Onishi}
\IEEEauthorblockA{\textit{Central Research Laboratories} \\
\textit{\tcr{NEC Corporation　AISTも追加}}\\
Kanagawa, Japan \\
t-onishi@bq.jp.nec.com}
}

\author{\IEEEauthorblockN{Shota Motoura\IEEEauthorrefmark{1}, Kazeto Yamamoto\IEEEauthorrefmark{1}\IEEEauthorrefmark{2}, Shumpei Kubosawa\IEEEauthorrefmark{1}\IEEEauthorrefmark{2}, Takashi Onishi\IEEEauthorrefmark{1}\IEEEauthorrefmark{2}}
\IEEEauthorblockA{\IEEEauthorrefmark{1} Central Research Laboratories, NEC Corporation,
Kanagawa, Japan\\
E-mail: \{motoura@ct, k-yamamoto@ft, kubosawa@cd, t-onishi@bq\}.jp.nec.com
\IEEEauthorblockA{\IEEEauthorrefmark{2}Artificial Intelligence Research Center, National Institute of Advanced Industrial Science and Technology, Tokyo, Japan\\
E-mail: \{kazeto.yamamoto, shumpei.kubosawa, takashi.onishi\}@aist.go.jp}
}

}

\maketitle

\begin{abstract}
This paper proposes a method to translate multilevel flow modeling (MFM) into a first-order language (FOL), which enables the utilisation of logical techniques, such as inference engines and abductive reasoners.
An example of this is a planning task for a toy plant that can be solved in FOL using abduction.
In addition, owing to the expressivity of FOL, the language is capable of describing actions and their preconditions.
This allows the derivation of procedures consisting of multiple actions.
\end{abstract}


\section{Introduction}
A plant is operated on the basis of its manual usually; however, it is not realistic that a manual contains instructions for all cases, especially regarding abnormal ones. 

For obtaining appropriate operation procedures for a wide variety of cases,  {\bf multilevel flow modeling} ({\bf MFM}) has been studied (\cite{b1,b2,b3}).
MFM is a functional modeling framework, in which a plant structure is expressed as a directed graph. 
The framework also has a set of {\bf influence propagation rules}, which consists of if-then rules regarding the states of related components.
If the state of a component has changed, the resulting state of the other components can  be obtained by applying the rules in the forward direction.
Conversely, given a desired state of a component, we can obtain the states of  other components to be satisfied for achieving the desired state by tracing {\it back} the propagation rules. 
This leads an action to a desired state.

Our contributions are as follows:
\begin{enumerate}
\item We propose a method to translate MFM into an FOL.
This enables the application of techniques used in the FOL to MFM, such as inference engines and abductive reasoners \cite{hobbs}.

\item Our method also enables MFM to employ planning techniques using abductive reasoners.
We give an example to illustrate that the planner can solve a planning task of a toy plant by using the translated MFM.
Moreover, since the FOL can express the preconditions for an action, 
operation procedures consisting of more than one action can automatically be derived, as is illustrated in Section \ref{application}.
\end{enumerate}

\section{Preliminaries}
\label{Preliminaries}
Before describing our method for translating MFM into an FOL, let us recall what MFM is.
{\bf Multilevel flow modeling （MFM）} is a functional modeling framework.
It models a plant as an augmented directed graph, whose vertices and edges are labelled with their {\bf function types} and {\bf relation types}, respectively. Such a model is called an {\bf MFM model} \cite{b3}.

\begin{figure}[h]
\begin{center}
  \includegraphics[width=8cm]{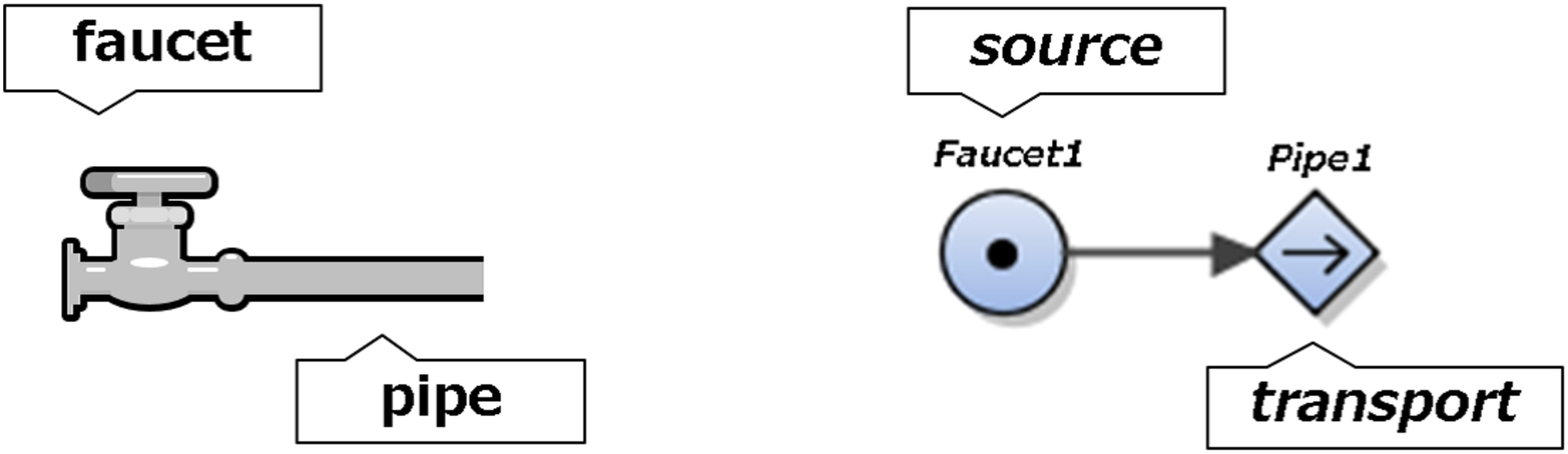}
  \caption{An actual structure (left) and its MFM model (right)}
  \label{fig: MFM example}
\end{center}
\end{figure}

\begin{figure}[t]
\begin{center}
  \includegraphics[width=8cm]{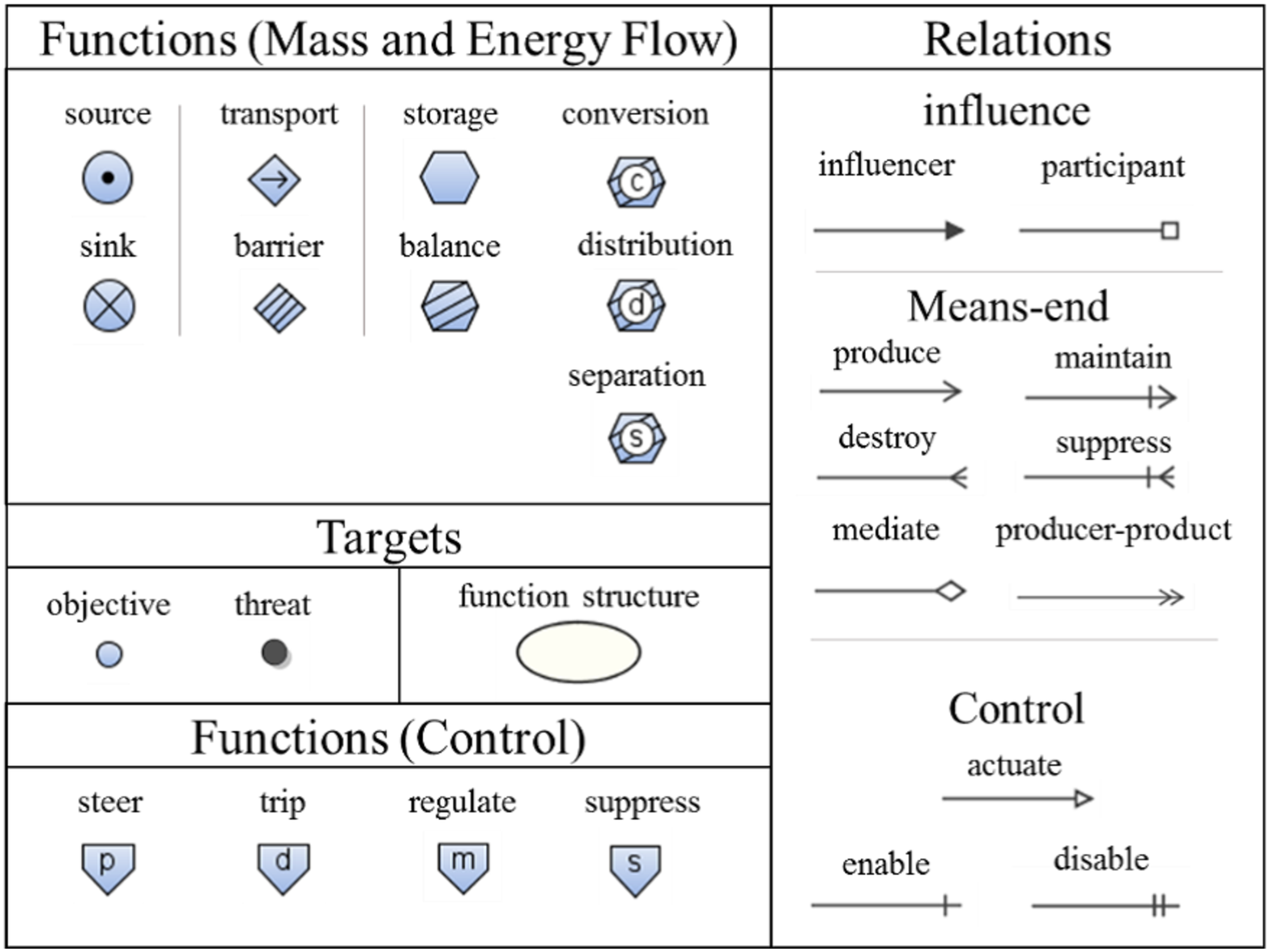}
\caption{MFM symbols \cite{b1}}
\label{fig: MFMsymbols}
\end{center}
\end{figure}

The example below illustrates the idea of MFM (Fig. \ref{fig: MFM example}).\footnote{For further details of MFM, refer to \cite{b3}.}
The structure in the left figure depicts a pipe connected to a faucet.
In MFM, this structure is modeled as the graph in the right figure.
Normal font and $italic$ $one$ are used to represent actual structures and MFM models, respectively.
Here, $Faucet1$ and $Pipe1$ are the names of the vertices,
which represent the faucet and the pipe in the left figure.
They are labelled with $\odot$ and $\Diamond$\hspace{-0.75em}\raisebox{0.5ex}{${\scriptscriptstyle \rightarrow}$}, which indicate the function types of {\it source} and {\it transport} (of water), respectively (see also Fig. \ref{fig: MFMsymbols} for all MFM symbols). 
The arrow ${\rightarrow}$ in the diamond $\Diamond$ expresses the direction of the {\bf flow} (of water).
A {\bf state} assigned to a vertex indicates 
the degree to which the component corresponding to the vertex satisfies the function corresponding to the function type.
For example, to a $source$ vertex, one of  $High/Low/No$ $Output$ $Flow$ is assigned; to the $transport$ vertex, one of  $High/Low/No$ $Flow$ is assigned.\footnote{Although $No$ $Output$ $Flow$ and $No$ $Flow$ may not be conventional, they are useful when giving an example of a procedure consisting of two consecutive actions in Section \ref{application}.}  (We often abbreviate 
``$Output$ $Flow$'' or ``$Flow$'' and simply say $High/Low/No$.)

\begin{figure}[h]
\begin{center}
  \includegraphics[width=8cm]{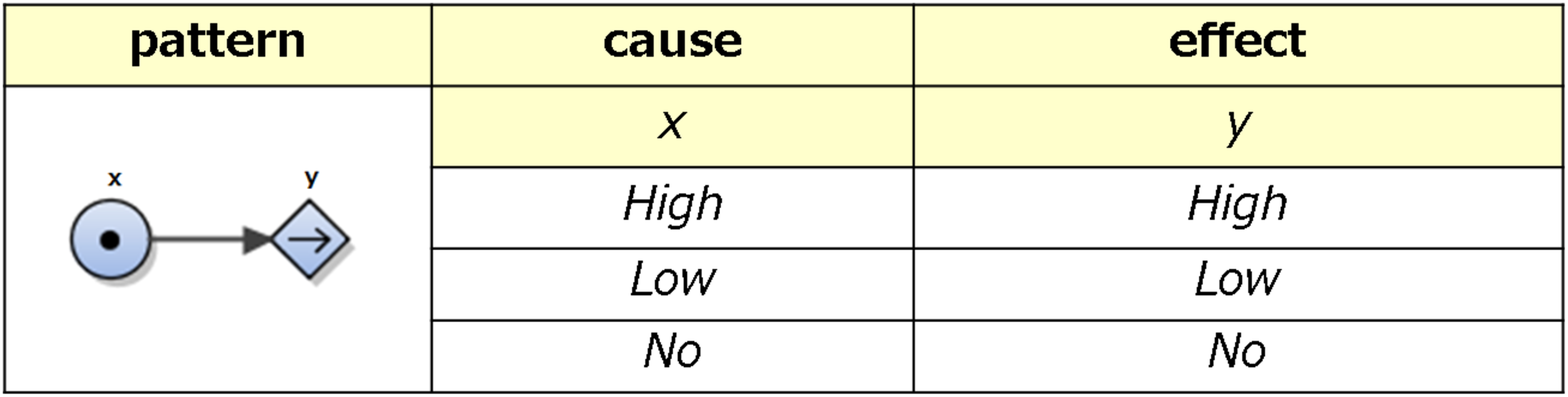}
\caption{Examples of influence propagation rules}
\label{fig: MFM rules}
\end{center}
\end{figure}

The arrow labelling the edge in the graph means that the state of $Faucet1$ affects that of $Pipe1$, as the inflow of the faucet affects the flow of the pipe.
The effect of a $source$ vertex on the related $transport$ vertex are pre-determined as an {\bf influence propagation rule}.
Fig. \ref{fig: MFM rules} shows the rules for the structure in Fig. \ref{fig: MFM example}.
A rule consists of three parts: a pattern, a cause and an effect (or effects when more than two vertices are involved).
Note that a rule, especially its pattern, is applicable regardless of the names of the vertices in the structure and therefore the names of vertices in the pattern are ``variables'' in this sense.

Given the state of a vertex, by tracing the influence propagation rules sequentially, the states of other vertices can be inferred; conversely, by tracing {\it back} the rules, the states of other vertices that lead to the given state of the vertex can be obtained. 
The study \cite{b1} uses this method to plan a plant operation action for a given target state of a vertex.

\section{The Target Language: FOL}\label{language}
The language which we use as the target language of our translation is a {\bf first-order language} ({\bf FOL}).\footnote{The general definitions of first-order language and first-order logic are not given here. Refer to \cite{b4} and\cite{b5} for them.}
One of the advantageous aspects of an FOL is its expressivity.
For example, an action and its preconditions can be expressed in the proposed language, as we see in Section \ref{application}.

Our language is defined in three stages: its terms; its atomic formulae; and its formulae.

A {\bf term} is a constant or a variable.
{\bf Constants} are ``names in {\it FOL}'' such as ${\tt Faucet1}$ and ${\tt Pipe1}$, which indicate the vertices whose names are $Faucet1$ and $Pipe1$ (${\tt typewriter}$ ${\tt font}$ is used to represent FOL).
{\bf Variables} are like ${\tt x}$ and ${\tt y}$, {\it etc.} They can be substituted with constants.
We capitalise the first letter of a constant and use a lower-case letter for a variable.

An {\bf atomic formula} is given as ${\tt p(t}_1{\tt ,\ldots,t}_n{\tt)}$, where $n\geq{0}$, ${\tt p}$ is a predicate symbol and ${\tt t}_i$ ($i=1,\ldots,n$) are terms.
The {\bf predicate symbols} we use are the following:
\begin{itemize}
\item ${\tt f}$ for each function type, $f$.\\
${\tt f(V)}$ means that vertex $V$ is labelled $f$.
For ${\tt source}$ function, for example, ${\tt source(Faucet1)}$ means vertex $Faucet1$ is labelled $source$.
\item
${\tt r}$ for each relation type, $r$.\\
${\tt r(V_1, V_2)}$ means that the edge from $V_1$ to $V_2$ is labelled $r$.
Thus, ${\tt influencer(Faucet1, Pipe1)}$ means that the edge from $Faucet1$ to $Pipe1$ is labelled $influencer$ .
\item ${\tt flow}$.\\ 
The meaning of ${\tt flow(V_1, V_2)}$ is that flow is going from $V_1$ to $V_2$; for example,
 ${\tt flow(Faucet1, Pipe1)}$ means that the flow is going from $Faucet1$ to $Pipe1$.
\item 
${\tt hold}$.\\
${\tt hold(V, S)}$ mean that vertex $V$ has the state $S$, so ${\tt hold(Faucet1, High)}$ is written to indicate that the state of $Faucet1$ is $High$. This predicate symbol is an analogue to ${\tt HoldsAt}$ in event calculus\cite{event}.
\end{itemize}

The {\bf formulae} are inductively constructed from atomic formulae by $\land$ (conjunction), $\lor$ (disjunction), $\neg$ (negation), and $\Rightarrow$ (implication).\footnote{The language does not use quantifiers, but we consider all variables in a formula are universally quantified.}
Thus, the followings are formulae:
\begin{center}
${\tt hold(x,High){\Rightarrow}hold(y,Low)}$;\\
${\tt hold(Faucet1, High){\lor}hold(Faucet1,Low)}$.
\end{center}

\section{Translation of MFM into FOL}\label{translation}
An MFM model and the influence propagation rules can be translated into the language we have defined above.

\subsection{MFM models}
Given an MFM model, we translate its vertices and its edges, separately.
For each vertex $V$:
\begin{itemize}
\item ${\tt f(V)}$ if the vertex is labelled with $f$,
\item ${\tt hold(V,S)}$ if the vertex has state $S$.
\end{itemize}
For each edge from $V_1$ to $V_2$:
\begin{itemize}
\item ${\tt r(V_1,V_2)}$ if the edge is labelled with $r$,
\item ${\tt flow(V_1,V_2)}$ if the flow is from $V_1$ to $V_2$.
\end{itemize}

Fig. \ref{fig: MFM example} illustrates this translation.

\begin{figure}[h]
\begin{center}
  \includegraphics[width=8cm]{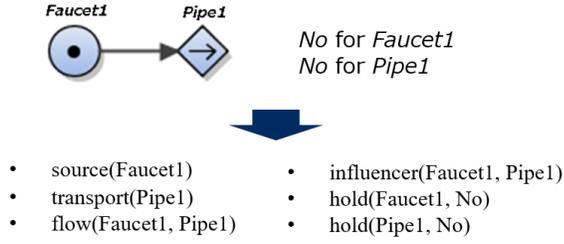}
  \caption{An MFM model with states (upper) and its FOL translation (lower)}
  \label{fig: translation}
\end{center}
\end{figure}

\subsection{Influence propagation rules}
As mentioned above, a rule is composed of three parts: a pattern, a cause and an effect. 
A pattern, $P$, can be considered as an MFM model without states.
Therefore, we translate it in a similar manner to the above and obtained (the conjunction of) the resulting formulae,  
$\mathcal{T}(P)$.
Note that, in an influence propagation rule, the pattern can be considered as a precondition.
The rest of the rules are its cause and its effect.
Their meaning can be written in the form that the state $S_1$ of vertex $x$ causes the state $S_2$ of vertex $y$.
This is translated into $${\tt hold(x, S_1){\Rightarrow}hold(y, S_2)}.$$
Here, ${\tt x}$ and ${\tt y}$ are variable and thus can be substituted with constants such as ${\tt Faucet1}$ and ${\tt Pipe1}$ (see also Section \ref{application}).
Combining these two, we obtain the translation of the propagation rule:
$$\mathcal{T}(P){\tt {\land}hold(x, S_1){\Rightarrow}hold(y, S_2)}.$$

For example, the rule that $High$ of $x$ implies $High$ of $y$ in Fig. \ref{fig: MFM rules} is translated as follows:
$$\mathcal{T}(P){\tt {\land}hold(x, High){\Rightarrow}hold(y, High)},$$
where $P$ is the pattern and $\mathcal{T}(P)$ is $${\tt source(x){\land}transport(y){\land}flow(x,y){\land}influencer(x,y)}.$$

\section{Application: Planning}\label{application}
Having introduced the translation of MFM into FOL, 
we can use techniques based on first-order logic, such as logical inference engines and abductive reasoners.
We here take the latter and illustrate that the translated MFM in FOL can be used to plan a plant operation procedure.

{\bf Abductive reasoning} is logical reasoning which uses if-then (i.e. implication) rules  in the {\it reverse} direction to obtain plausible hypotheses.
Abduction is known to be applicable to solve a planning task \cite{shanahan}.
As mentioned before, our method can automatically derive a procedure which consists of more than one action, because our language is expressive enough to describe actions and their preconditions.

\subsection{Example 1: single action}
Let us consider Fig. \ref{fig: MFM example} and suppose that the faucet is closed and, thus, there is no water flow in the pipe.
This is translated as in Fig. \ref{fig: translation}.
In addition, we also suppose that the faucet can be opened, which can also be written in our FOL as follows:
$${\tt open(Faucet1){\land}hold(Faucet1, No){\Rightarrow}hold(Faucet1, High)}.$$
In this setting, we consider a planning task to change the current state ${\tt hold(Faucet1, No)\land{hold}(Pipe1, No)}$ to the target state ${\tt hold(Faucet1,High)}$. A correct plan is opening the faucet, which can be derived automatically.

\begin{figure*}
\begin{center}
  \includegraphics[width=16cm, height=6.5cm]{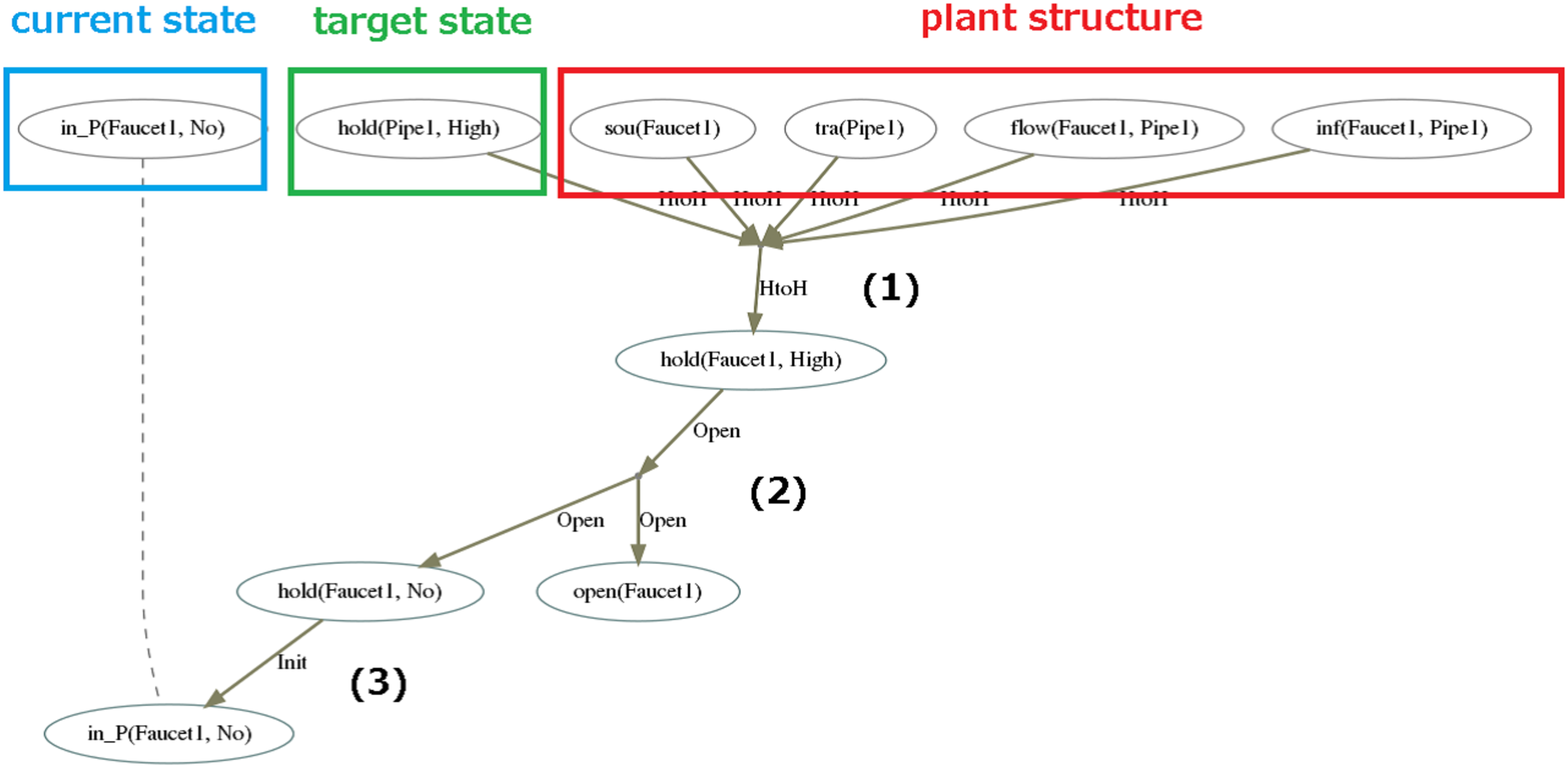}
\caption{Planning of a single action: the dotted line indicates the two vertices are the same, and in\_P means that the state is positive at the initial time, which is a practical technique and not essential.}
\label{fig: plan1}
\end{center}
\end{figure*}

Figure \ref{fig: plan1} shows the planned procedure and inference rules used during the planning:
each vertex represents an atomic formula; and each hyper edge (i.e. a fork-shaped ``edge'' in the figure between two sets of vertices) represents an implication. Note that the direction of an arrow is reversed, since this planning method employs abduction.
The planning task is solved backwards from the target state ${\tt hold(Faucet1,High)}$.
(1) We first consider the following instance of an influence propagation rule with ${\tt x}={\tt Faucet1}$ and ${\tt y}={\tt Pipe1}$:${}$
\begin{flushleft}
${\tt source(Faucet1){\land}transport(Pipe1)\land}$\\
${\tt {\land}flow(Faucet1,Pipe1)\land{influencer(Faucet1,Pipe1)}\land}$
\\${\tt {\land}hold(Faucet1, High)}{\tt {\Rightarrow}hold(Pipe1, High)}$.
\end{flushleft}
Since the plant structure part, i.e. the first four literals, is already satisfied,
it is enough to obtain ${\tt hold(Faucet1, High)}$ to achieve ${\tt hold(Pipe1, High)}$.
(2) We then consider the rule given above for action ${\tt open(Faucet1)}$.
Applying this rule in the reverse direction, we see that, to achieve ${\tt hold(Faucet1, High)}$, an action of ${\tt open(Faucet1)}$ on the state ${\tt hold(Faucet1, High)}$ suffices.
(3)  This state is the current state and thus already satisfied.

As described above, we obtain the procedure consisting of an action ${\tt open(Faucet1)}$, which is a correct answer as we mentioned.

\subsection{Example 2: multiple actions}\label{example2}
\begin{figure*}
\begin{center}
  \includegraphics[width=16cm, height=8cm]{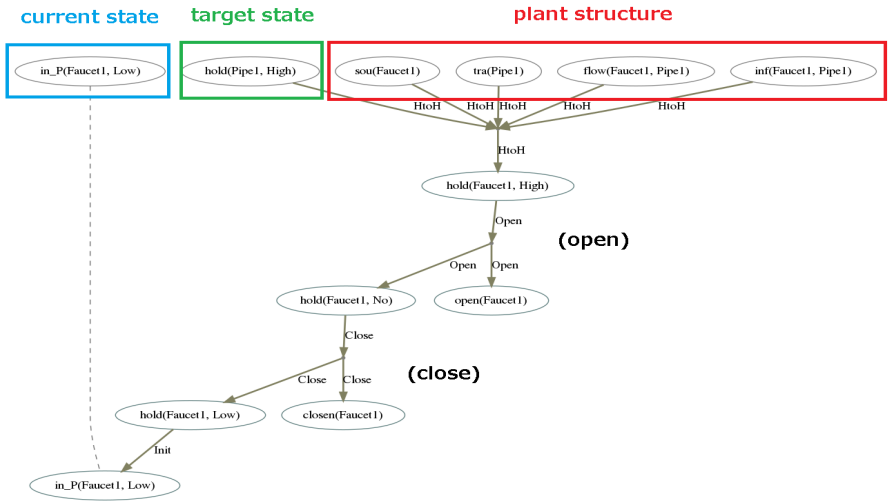}
  \caption{Planning of a procedure consisting of multiple actions}
  \label{fig: plan2}
\end{center}
\end{figure*}

Next, a condition is imposed upon the setting above: the faucet can be opened only when it is closed.
More precisely, we consider the following two actions:${}$
\begin{flushleft}
${\tt open(Faucet1){\land}hold(Faucet1, No){\Rightarrow}hold(Faucet1, High)}$;\\
${\tt close(Faucet1){\land}hold(Faucet1, Low){\Rightarrow}hold(Faucet1, No)}.$
\end{flushleft}
Again, the target state is that the flow in the pipe be high, ${\tt hold(Faucet1,High)}$, while the current state is that the faucet is half-open; thus, the flow in the pipe is low:
${\tt hold(Faucet1, Low)\land{hold}(Pipe1, Low)}$.
A correct plan is to closed the faucet completely at first, and then fully open it.\
This can be derived in a similar way to the above, and Fig. \ref{fig: plan2} shows the result of automatic planning.

\section{Summary and Future Work}
In this paper, we proposed a method for translating MFM into FOL (Section \ref{translation}) and a method for planning plant operation procedures and an application of translated MFM (Section \ref{application}).
We saw that, by following our method, plans consisting of not only one action but also multiple actions can be derived automatically (Section \ref{application}).

We shall address the following issues in future work:
\begin{itemize}
\item {\it Introduction of the concept of time.}
A procedure planned by our proposed method is not a set of formulae but a graph.
Therefore, we cannot apply logical techniques to {\it plans}.
An approach to tackle this problem is to introduce the concept of time to our model and 
modify the planning method to describe the partial order of actions, which is currently represented as a graph, explicitly as formulae.

\item {\it Translation from P\&ID into FOL.} Pipe and instrument diagram (P\&ID) is a widely-used plant representation framework. 
If a given plant structure represented in P\&ID can automatically be translated into FOL via MFM, then the automated plant operation planning technique given in this paper will be applicable at low cost.
\end{itemize}

\section*{Acknowledgment}
The authors sincerely appreciate Professor Akio Gofuku for his comments and advice on the research.

\end{document}